\documentclass[11pt]{article}

\usepackage[preprint]{acl}
\usepackage{fontawesome5} 
\usepackage{times}
\usepackage{latexsym}

\usepackage[T1]{fontenc}

\usepackage[utf8]{inputenc}

\usepackage{microtype}

\usepackage{inconsolata}

\usepackage{graphicx}

\title{OpenHospital: A Thing-in-itself Arena for Evolving and Benchmarking LLM-based Collective Intelligence}

\author{
  \textbf{Peigen Liu}\textsuperscript{1}, 
  \textbf{Rui Ding}\textsuperscript{1}, 
  \textbf{Yuren Mao}\textsuperscript{1}\thanks{Corresponding author.}, 
  \textbf{Ziyan Jiang}\textsuperscript{1}, 
  \textbf{Yuxiang Ye}\textsuperscript{1}, 
  \textbf{Yunjun Gao}\textsuperscript{1},\\
  \textbf{Ying Zhang}\textsuperscript{2},
  \textbf{Renjie Sun}\textsuperscript{2},
  \textbf{Longbin Lai}\textsuperscript{3},
  \textbf{Zhengping Qian}\textsuperscript{4} \\
  \textsuperscript{1}School of Software Technology, Zhejiang University \\
  \textsuperscript{2}Laboratory for Statistical Monitoring and Intelligent Governance of Common Prosperity, \\Zhejiang Gongshang University \\
  \textsuperscript{3}Tongyi Lab, Alibaba Group 
  \textsuperscript{4}Alibaba Cloud \\
  \texttt{\{peigenliu, rui.ding, yuren.mao\}@zju.edu.cn} \\
\raisebox{-0.2ex}{\faGithub} \hspace{0.5em} \href{https://github.com/ZJU-LLMs/Agent-Kernel}{https://github.com/ZJU-LLMs/Agent-Kernel}
}

\begin{document}
\maketitle
\raggedbottom
\begin{abstract}

Large Language Model (LLM)-based Collective Intelligence (CI) presents a promising approach to overcoming the data wall and continuously boosting the capabilities of LLM agents. However, there is currently no dedicated arena for evolving and benchmarking LLM-based CI. To address this gap, we introduce OpenHospital, an interactive arena where physician agents can evolve CI through interactions with patient agents. This arena employs a data-in-agent-self paradigm that rapidly enhances agent capabilities and provides robust evaluation metrics for benchmarking both medical proficiency and system efficiency. Experiments demonstrate the effectiveness of OpenHospital in both fostering and quantifying CI.

\end{abstract}

\section{Introduction}

\centerline{\textit{Thing in itself}}

\rightline{---Immanuel Kant}

The scaling laws governing Large Language Models (LLMs) currently encounter significant limitations as high-quality human corpora reach exhaustion, termed the \textit{data wall}. Overcoming this barrier requires Collective Intelligence (CI), wherein multi-agent coordination facilitates continuous interaction data synthesis. However, transitioning toward CI demands a paradigm shift in data methodology: progressing from training agents on static, invariant inputs to embedding them within dynamic environments that iteratively evolve based upon complex autonomous agent behaviors.


This shift echoes a fundamental Kantian dichotomy between the \textit{phenomenon}—the object as perceived—and the \textit{noumenon}, or the \textit{thing-in-itself (das Ding an sich)} \citep{kant1781critique}. 
Existing static datasets capture only the phenomenon, restricting agents to a superficial level of imitation and rote memorization. For true collective intelligence to emerge, agents must transcend surface-level observations and approximate the noumenal structure of reality through active, purposeful engagement. Current research, however, remains largely confined to static benchmarks that evaluate snapshot capabilities, lacking the dynamic arena required for the evolution of collective cognition.


To bridge this gap, we propose \textbf{OpenHospital}, a thing-in-itself arena for the evolving and benchmarking of LLM-based collective intelligence. OpenHospital introduces a novel data paradigm termed \textbf{data-in-agent-self}: rather than receiving static inputs (phenomena), physician agents must interact with patient agents—modeled as dynamic entities (noumena)—to elicit clinical information. This interaction forces physicians to integrate medical knowledge and debate treatment options, driving the emergence of collective intelligence. To function as a valid training ground for such intelligence, OpenHospital satisfies three critical requirements: (i) \textbf{Quantifiable Evolution}, utilizing rigorous metrics such as Examination Precision, Diagnostic Accuracy, and Treatment Plan Alignment to track the emergence of intelligence; (ii) \textbf{Dynamic Complexity}, creating a non-deterministic environment where patient responses unfold dynamically; and (iii) \textbf{Collaborative Necessity}, establishing scenarios where rare diseases and comorbidities make agent cooperation indispensable.

To validate the effectiveness of OpenHospital,
we establish a baseline for it using the Agent-Kernel framework \citep{mao2025agentkernelmicrokernelmultiagentframework}. Our experiments demonstrate that as case volume increases, physician agents exhibit significant improvements across all clinical capability metrics. Furthermore, analysis of the diagnostic process reveals the spontaneous emergence of sophisticated cooperative behaviors, including peer consultation and consensus-driven decision-making. These results validate that OpenHospital functions not only as a rigorous evaluation benchmark for agents in changing environments but also as a potent evolutionary training ground for genuine collective intelligence.

\section{Related Work}
\paragraph{LLM-based Multi-Agent Systems}
With the development of large language models (LLMs), LLM-based multi-agent systems (LLM-MAS) have progressed, contributing to the study of collective intelligence. Existing works generally fall into two categories: Task-Oriented and Simulation-Oriented. In task-oriented domains, LLM-based multi-agent systems are applied to fields like software engineering\citep{hong2024metagpt,wu2023autogenenablingnextgenllm,qian2023chatdev}, scientific research\citep{gottweis2025aicoscientist,ghafarollahi2024sciagentsautomatingscientificdiscovery,su2025headsbetteroneimproved}, and embodied AI\citep{mandi2023rocodialecticmultirobotcollaboration,yu2025conavgptmultirobotcooperativevisual}. In simulation-oriented tasks, following the pioneering work of Stanford Smallville\citep{park2023generativeagentsinteractivesimulacra}, many studies have emerged in fields like social\citep{park2023generativeagentsinteractivesimulacra,mao2025agentkernelmicrokernelmultiagentframework,piao2025agentsocietylargescalesimulationllmdriven}, narrative\citep{huot2025agentsroomnarrativegeneration, yu-etal-2025-multi}, economic\citep{li2024econagentlargelanguagemodelempowered,lin2025simulatingmacroeconomicexpectationsusing,sashihara2025llmbasedmultiagentsimulatingstrategic} and other professional simulations\citep{zhang-etal-2025-simulating, bougie2025citysimmodelingurbanbehaviors}. While these works emphasize the construction of robust static MAS, they provide a limited perspective on the dynamic processes, which enable the system to iteratively refine agent behaviors and achieve emergent intelligence. To address this issue, this work introduces a training ground that is measurably evolvable, dynamically complex, and inherently collaborative, specifically designed to drive agent evolution.

\paragraph{Benchmarks for LLM-based Multi-Agent Systems}
As MAS research grows, many benchmarks have emerged to evaluate their performance. These benchmarks use various scenarios and metrics and generally fall into two categories: Role-playing and Task-driven. Role-playing focuses on social interactions, and Task-driven focuses on task completion. In role-playing, benchmarks cover fields like social simulation\citep{zhou2024sotopiainteractiveevaluationsocial,park2023generativeagentsinteractivesimulacra, piao2025agentsocietylargescalesimulationllmdriven}, competition and cooperation\citep{zhu2025multiagentbenchevaluatingcollaborationcompetition}. In task-driven scenarios, benchmarks focus on specific applications like software engineering\citep{jimenez2024swebenchlanguagemodelsresolve,golnari2026devbenchrealisticdeveloperinformedbenchmark}, scientific research\citep{chen2025scienceagentbenchrigorousassessmentlanguage}, and mathematical reasoning\citep{gao2024omnimathuniversalolympiadlevel}. However, some benchmarks rely on static datasets. As a result, they cannot evaluate how agents react in dynamic environments. Although some studies have started to address this issue, their evaluation processes are still limited, relying heavily on subjective LLM-based scoring or involving only highly restricted, human-defined dynamics. Furthermore, their assessment metrics are not comprehensive enough. They validate the entire process based only on the final result, which is not enough for complex simulations. To bridge this gap, we adopt clinical diagnosis---a knowledge-intensive, collaborative, and objectively evaluable scenario—as a dynamic testbed for tracking agent evolution.

\paragraph{Multi-Agent Systems for Healthcare}
Recent research has increasingly focused on simulating clinical diagnosis scenarios. For instance, Agent Hospital\citep{li2025agenthospitalsimulacrumhospital} pioneered the closed-loop simulation of real hospital workflows and has been deployed. While Agent Clinic\citep{schmidgall2025agentclinicmultimodalagentbenchmark} focuses on performance evaluation in complex environments. MedAgentSim\citep{almansooriandkumarMedAgentSim} uses features like multi-turn doctor-patient dialogues to create more realistic simulations. However, most previous studies focused on "textbook" cases with clear symptoms and easy diagnoses. To address this, we created a new dataset featuring comorbidities and long-tail diseases to increase realism and challenge for agents. These cases are highly complex and often require consultations across different medical departments. Therefore, unlike previous work, our approach encourages frequent collaboration and consultation between agents.

\section{Construction of Patient Agents}

To effectively stimulate LLM-based collective intelligence within the OpenHospital arena, the construction of patient agents requires a rigorous balance between clinical validity and realistic human variability. We define and implement four indispensable pillars for realistic patients simulation in healthcare: \textbf{Clinical Correctness}, \textbf{Persona Diversity}, \textbf{Linguistic Fluency}, and \textbf{Behavioral Realism}. This section details the systematic construction of these agents and the validation metrics employed to verify their practical effectiveness.

\subsection{Clinical Correctness}

Clinical correctness mandates that patient agents strictly adhere to established medical knowledge, ensuring that synthesized symptoms, disease progressions, and clinical histories are logically coherent and physiologically plausible.

To achieve such medical rigor, we develop a multi-stage data synthesis pipeline powered by DeepSeek-v3.1 model \citep{deepseek2024v3}. The process begins with the curation of an authoritative knowledge base encompassing 583 distinct diseases and 467 binary comorbidities across 19 clinical departments. Utilizing these resources, the pipeline synthesizes comprehensive patient data comprising three core components: \textit{patient ontology}, \textit{examination reports}, and \textit{ground-truth diagnoses}. Here, the ontology serves as a structured domain conceptualization \citep{ghanadbashi2024ontologyenhanced}, incorporating real-world clinical schemas to represent both disease dimensions and persona attributes. Within the synthesis stage, to address the hallucinations inherent in single-pass generation, we implement a multi-step refinement framework. Rather than generating records in a single execution, this framework iteratively models patient attributes under epidemiological constraints, strictly enforcing logical consistency between presenting symptoms, diagnostic findings, and final diagnoses to guarantee medical plausibility.

To validate clinical correctness, we design the targeted evaluation of the agents' data foundations:
\begin{itemize}
  \item \textbf{Q: How clinically reliable is the static data backing the agents?} We utilize an LLM-as-a-judge system (powered by GPT-5.2 \citep{openai2025gpt52}) to assess the internal consistency of the synthesized records. The evaluation focuses on the causal alignment between symptoms and diagnoses, the temporal progression of patient histories, and the safety of implied clinical pathways. As shown in Table \ref{tab:patient_data_evaluation}, the static data achieve an average medical consistency score of \textbf{4.4113} (on a 1–5 scale), confirming a robust internal logic that respects medical common sense.
\end{itemize}

\begin{table}[ht]
  \centering
  \begin{tabular}{lll}
    \hline
    \textbf{Metric} & \textbf{Score} \\
    \hline
    Self-BLEU4 & 0.4111 \\
    TF-IDF Diversity & 0.8727 \\
    [1ex] 
    Perplexity & 5.7501 \\
    [1ex] 
    Medical Consistency & 4.4113 \\
    \hline
  \end{tabular}
  \caption{\label{tab:patient_data_evaluation}
    Evaluation results of the synthesized static patient data. 
  }
\end{table}

\subsection{Persona Diversity}
Persona diversity is defined as the inherent variance in demographic backgrounds, psychological traits, and behavioral patterns among the simulated patient population, ensuring a heterogeneous and realistic environment for robust agent evaluation.

To realize this diversity, we construct a comprehensive persona seed library by extracting and cleaning non-sensitive attributes related to personalities. These structured attributes serve as controllable inputs alongside the medical knowledge base, which ensures that the synthesized patient records maintain reasonable diversity. During the agent role-playing phase, these specific demographic and psychosocial characteristics are dynamically injected via prompt engineering to natively modulate the agents' tone, vocabulary, and conversational style. This deliberate integration prevents the patient agents from defaulting to the repetitive, template-like communication patterns commonly observed in standard LLM simulations.

\begin{figure}[ht]
\includegraphics[width=\columnwidth]{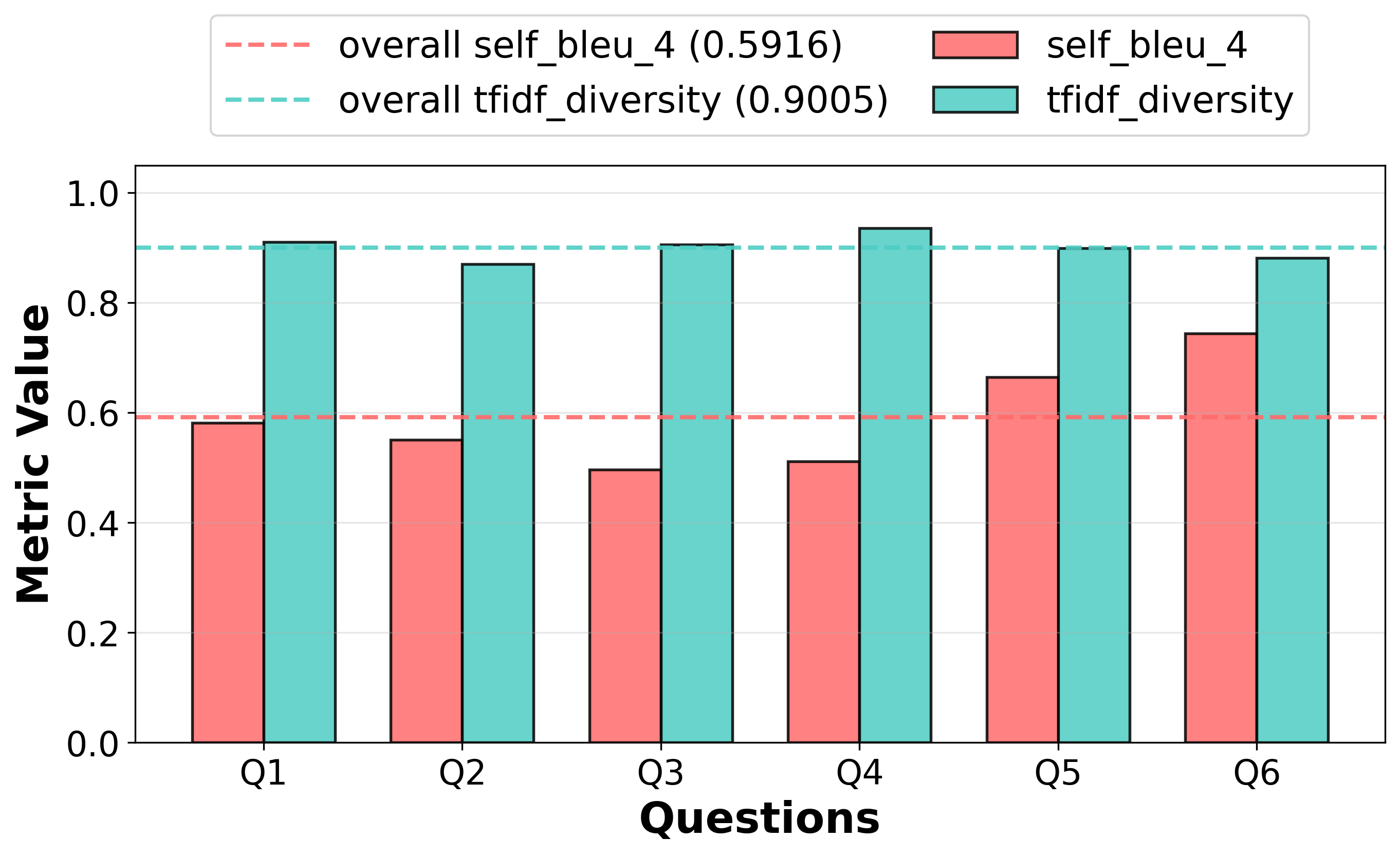}
\caption{Interactional response diversity of patient agents across the fixed question set.}
\label{fig:response_diversity}
\end{figure}

To quantify the effectiveness of our persona modeling, we evaluate the distinctness of both the static profiles and the dynamic conversational outputs:
\begin{itemize}
    \item \textbf{Q: Does the synthesized static dataset provide sufficient demographic and identity variance?} We utilize Self-BLEU4 and TF-IDF diversity metrics to quantify the distinctness of the generated patient personas. As shown in Table \ref{tab:patient_data_evaluation}, the static dataset attains an average Self-BLEU4 score of \textbf{0.4111} and a TF-IDF diversity score of \textbf{0.8727}. These metrics indicate substantial variance in patient identities, confirming that our seed library successfully avoids profile collapse and ensures a broad representational spectrum for evolving.
    \item \textbf{Q: Do the dynamic patient agents react differently to the same situation?} We assess interactional diversity by subjecting all patient agents to a fixed set of standardized clinical inquiries. By calculating diversity metrics grouped by these specific questions, we find that the dynamic responses achieve an average Self-BLEU4 of \textbf{0.5916} and a TF-IDF diversity score of \textbf{0.9005} (see Figure \ref{fig:response_diversity}). These results validate that identical clinical inputs yield naturally persona-modulated responses, accurately reflecting the behavioral variance observed in real-world patients.
\end{itemize}

\subsection{Linguistic Fluency}
Linguistic fluency in our work is defined as the lexical richness and semantic coherence of the generated clinical narratives, ensuring the medical records produced are both professional and non-stereotypical.

To achieve this, our synthesis pipeline leverages the advanced natural language generation capabilities of DeepSeek-v3.1, guided by persona-specific prompts and grounded in curated medical encyclopedias to ensure accurate nomenclature. Rather than merely optimizing for syntactic predictability, this approach emphasizes the production of diverse, high-fidelity clinical records that reflect the nuances of a real-world patient population.

To evaluate the linguistic quality of the synthesized data, we measure its distributional alignment with professional medical corpora:
\begin{itemize}
    \item \textbf{Q: How natural and coherent is the medical text across these distinct personas?} We compute the Perplexity (PPL) across the entire dataset using the Baichuan-M2-32B \citep{baichuan2025m2}, a domain-specialized medical model, as an independent evaluator. In this context, a lower PPL indicates that the synthetic records align well with the distribution of professional clinical texts. The generated records yield an overall perplexity score of \textbf{5.7501}, demonstrating high linguistic coherence and strong alignment with standard clinical terminology and grammatical conventions.
\end{itemize}

\subsection{Behavioral Realism}
Behavioral realism denotes the capacity of patient agents to emulate the authentic cognitive and interactional constraints of real-world clinical consultations.

To implement this characteristic, we employ a dynamic role-playing architecture designed to mimic actual clinical environments. We establish a strict epistemic boundary of information asymmetry: the patient agent only accesses subjective ontological information (e.g., experienced symptoms and personal history), while objective diagnoses and examination reports remain hidden. Specifically, synthesized medical histories are stored in a vector database to serve as the agent's semantic memory, which is retrieved organically during the conversation. Crucially, agents are explicitly instructed to disclose information only in response to targeted inquiries. This constraint prevents unrealistic information dumping, thereby compelling the physician agent to demonstrate active and logical diagnostic reasoning.

\begin{figure}[ht]
\includegraphics[width=\columnwidth]{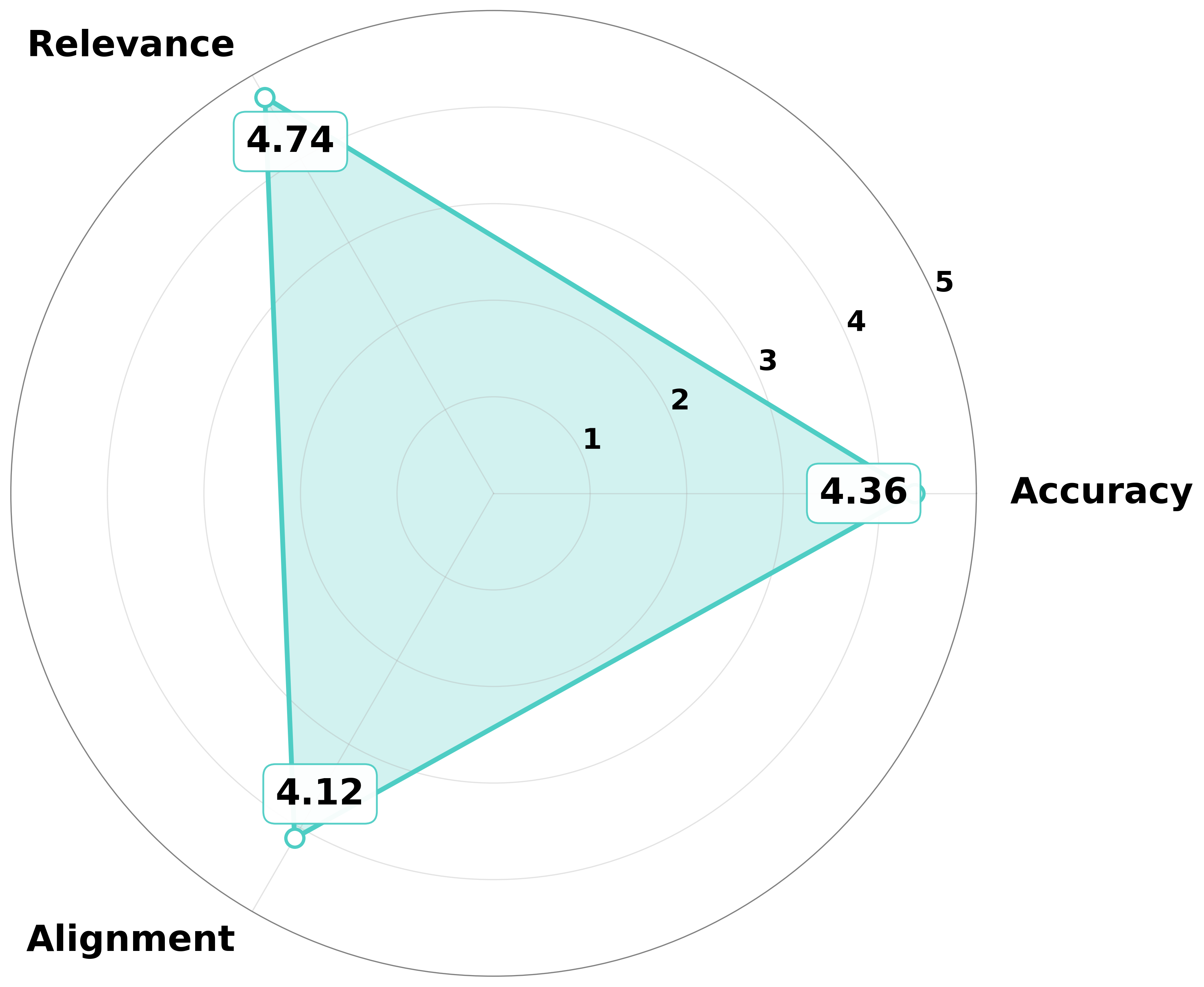}
\caption{Dynamic response quality scores evaluated by GPT-5.2 across Accuracy, Relevance, and Persona Alignment.}
\label{fig:response_quality}
\end{figure}

To validate the interactional fidelity of this architecture, we assess agent performance within a simulated multi-turn consultation framework:
\begin{itemize}
    \item \textbf{Q: How effectively do agents maintain character and context during complex, multi-turn clinical dialogues?} We evaluate dynamic response quality using GPT-5.2 as an expert judge, providing it with complete patient data to assess the hidden states of the interactions. Across simulated, non-linear consultations driven by physician inquiries, the patient agents achieve high average scores: \textbf{4.36/5} for Accuracy, \textbf{4.74/5} for Relevance, and \textbf{4.12/5} for Persona Alignment (as illustrated in Figure \ref{fig:response_quality}). These results confirm that the agents successfully navigate evolving clinical contexts, maintaining strict behavioral fidelity and authentic information-sharing pacing without breaking character.
\end{itemize}

\section{Benchmarking Agent Evolution}

To validate the efficacy of the OpenHospital arena in fostering collective intelligence, we establish a comprehensive baseline system. In this section, we first introduce a suite of multi-dimensional metrics designed to capture both clinical validity and systemic efficiency. We then deploy our baseline to evaluate these metrics, demonstrating the continuous evolution of physician agents.

\subsection{Evaluation Metrics}
We propose a multi-dimensional evaluation comprising Medical Capability, which evaluates clinical validity from examinations to treatment, and System Efficiency, which assesses computational cost.

\paragraph{Medical Capability Metrics} 
We evaluate clinical proficiency using three complementary metrics:
\begin{itemize}
    \item \textbf{Examination Precision} assesses the relevance and necessity of ordered tests. Defined as $ P = |L_{pred} \cap L_{gold}| / |L_{pred}| $, where $L_{pred}$ is the predicted examination list and $L_{gold}$ is the standard clinical set, this metric penalizes unnecessary tests while rewarding the prioritization of informative examinations.
    \item \textbf{Diagnostic Accuracy} measures the correctness of the final consensus diagnosis. Formally, for a case $i$ with ground truth $D_{gold}$, the score is $1$ if the agent's diagnosis $D_{agent} = D_{gold}$, and $0$ otherwise, capturing the core clinical judgment in dynamic and incomplete information environments.
    \item \textbf{Treatment Plan Alignment} evaluates therapeutic quality against gold-standard guidelines via an LLM-based evaluator. Plans are scored across three dimensions—safety, effectiveness, and personalization—providing a holistic view of the agents' therapeutic reasoning capabilities.
\end{itemize}

\paragraph{System Efficiency Metric} Beyond medical capability, we assess the computational cost of system execution using \textbf{Total Input Tokens}. This metric measures the cumulative number of input tokens processed across all LLM interactions, including patient simulations, physician reasoning, and auxiliary
evaluations, over the full execution workflow. As an end-to-end indicator of prompt-side computational demand, it provides a practical basis for analyzing the trade-off between task performance and computational efficiency in real-world deployment settings.

The proposed evaluation metrics balance clinical validity with systemic efficiency to ensure a robust assessment of agent evolution. By integrating metrics across the full clinical pathway—examination, diagnosis, and treatment—we transcend the limitations of single-dimension accuracy and prevent shortcut reasoning, where correct diagnoses might stem from flawed investigative processes or unsafe therapeutic plans. Simultaneously, Total Input Tokens serves as a critical proxy for computational overhead and operational cost. Tracking this metric ensures that performance gains reflect optimized reasoning logic rather than exhaustive, inefficient prompting. Together, these multi-dimensional indicators provide a rigorous benchmark for the scalability and cost-effectiveness of collective intelligence in real-world healthcare deployments.

\subsection{Experimental Setup}
\paragraph{Datasets}
The experiment is conducted on a comprehensive dataset comprising 12,000 diverse patient records, which is partitioned into training and test sets at a 9:1 ratio. To meticulously track the evolutionary trajectory of the system’s collective intelligence, the training set is further organized into 22 sequential batches, with each batch containing approximately 500 clinical cases. 

\begin{figure*}[t]
    \centering
    \includegraphics[width=\linewidth]{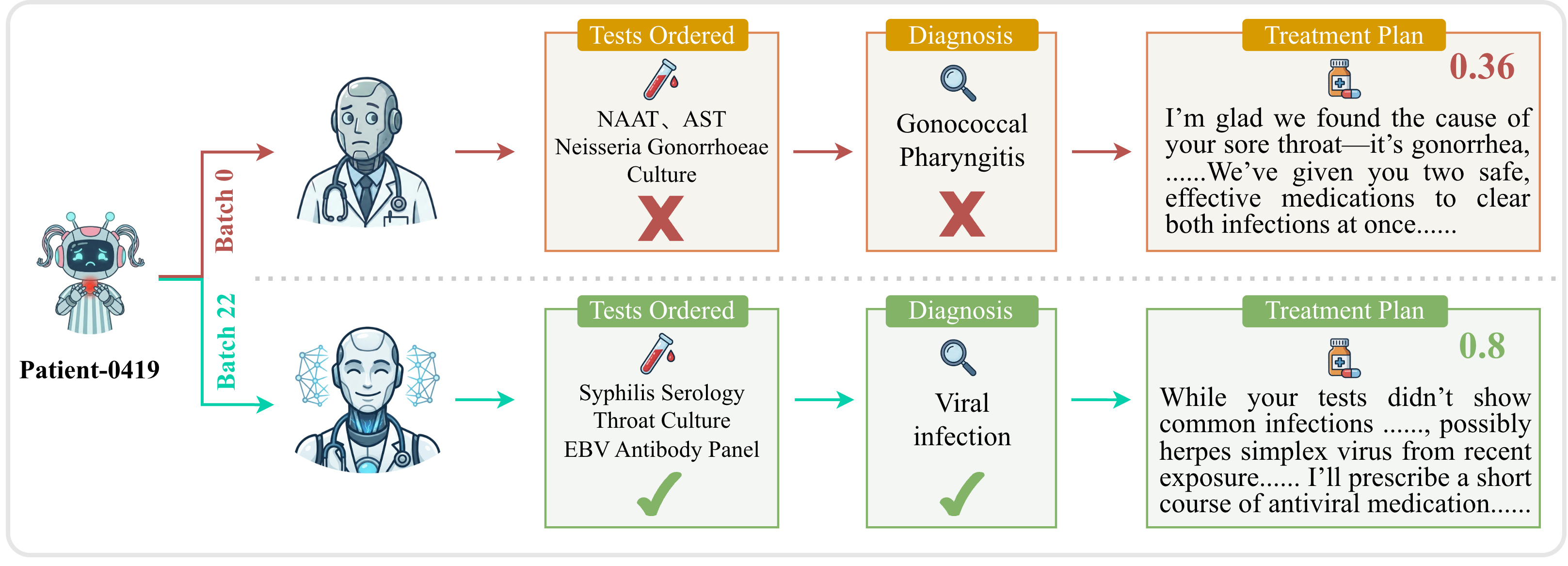}
    \caption{Diagnostic evolution of a physician agent across training batches.}
    \label{fig:evolution_case}
\end{figure*}

\paragraph{Baseline}
Our baseline is constructed utilizing the Agent-Kernel framework, featuring a multi-agent architecture of 38 physician agents distributed across 19 specialized clinical departments (two physicians per department). These agents operate within a sophisticated action space that encompasses patient perception, targeted inquiry, diagnostic examination, multi-agent consultation, and knowledge retrieval, culminating in final treatment formulations. Central to this baseline is a closed-loop reflection mechanism designed to drive autonomous evolution; after each case, agents engage in a multi-dimensional self-critique that synthesizes diagnostic accuracy against ground truth, examination efficiency, and therapeutic safety to bridge efficacy gaps. By integrating these diagnostic, investigative, and treatment reflections into a unified feedback loop, the agents systematically accumulate clinical experience and optimize their decision-making logic over time.

\paragraph{Experimental Settings}
For the experimental implementation, both physician and patient agents are powered by the Qwen/Qwen3-Next-80B-A3B-Instruct model \citep{qwen2024qwen3} to ensure high-fidelity interaction and reasoning.

\subsection{Evaluation Results of Agent Evolution}

To robustly validate OpenHospital's efficacy as an evolutionary arena, we conduct a comprehensive quantitative analysis of the physician agents across 22 iterative training batches. We evaluate this evolution through two distinct but complementary lenses: the enhancement of clinical proficiency and the optimization of computational efficiency. 

\begin{figure}[ht]
\includegraphics[width=\columnwidth]{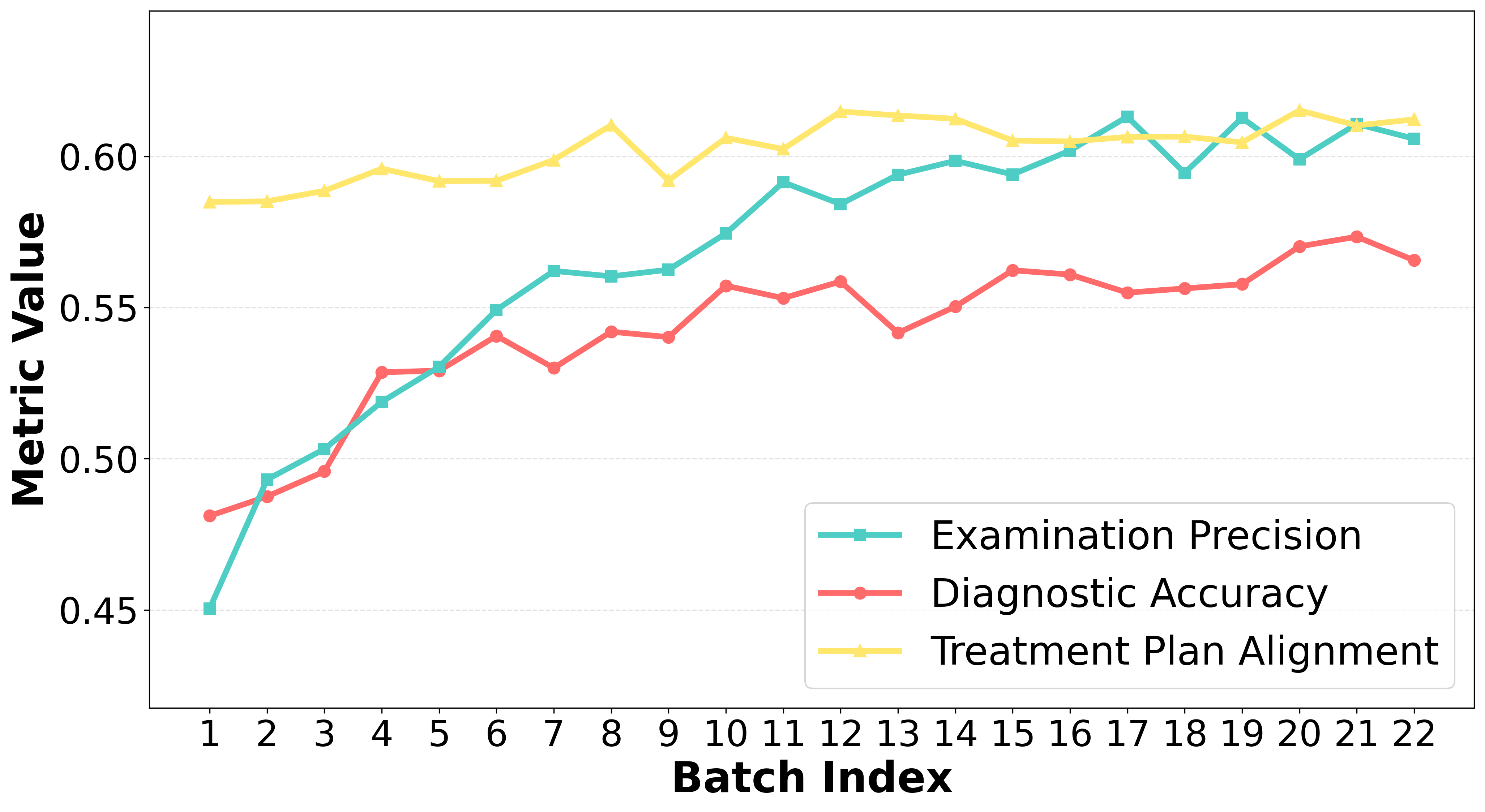}
\caption{Performance trajectories of clinical capabilities across training batches.}
\label{fig:medical_capabilities_trend}
\end{figure}

\paragraph{Medical Capabilities} We track the performance trajectories of the physician agents throughout the training timeline. As Figure \ref{fig:medical_capabilities_trend} illustrates, the agents exhibit consistent upward trends across all core clinical metrics as they accumulate experience. Most notably, Examination Precision shows a substantial gain, rising from \textbf{45.05\%} to \textbf{61.31\%}, indicating a marked reduction in exploratory or redundant testing. Concurrently, Diagnostic Accuracy demonstrates strong improvement, climbing from \textbf{48.11\%} to \textbf{57.34\%}, while Treatment Plan Alignment shows a steady and consistent refinement, increasing from \textbf{58.49\%} to \textbf{61.52\%}.

\begin{figure}[ht]
\includegraphics[width=\columnwidth]{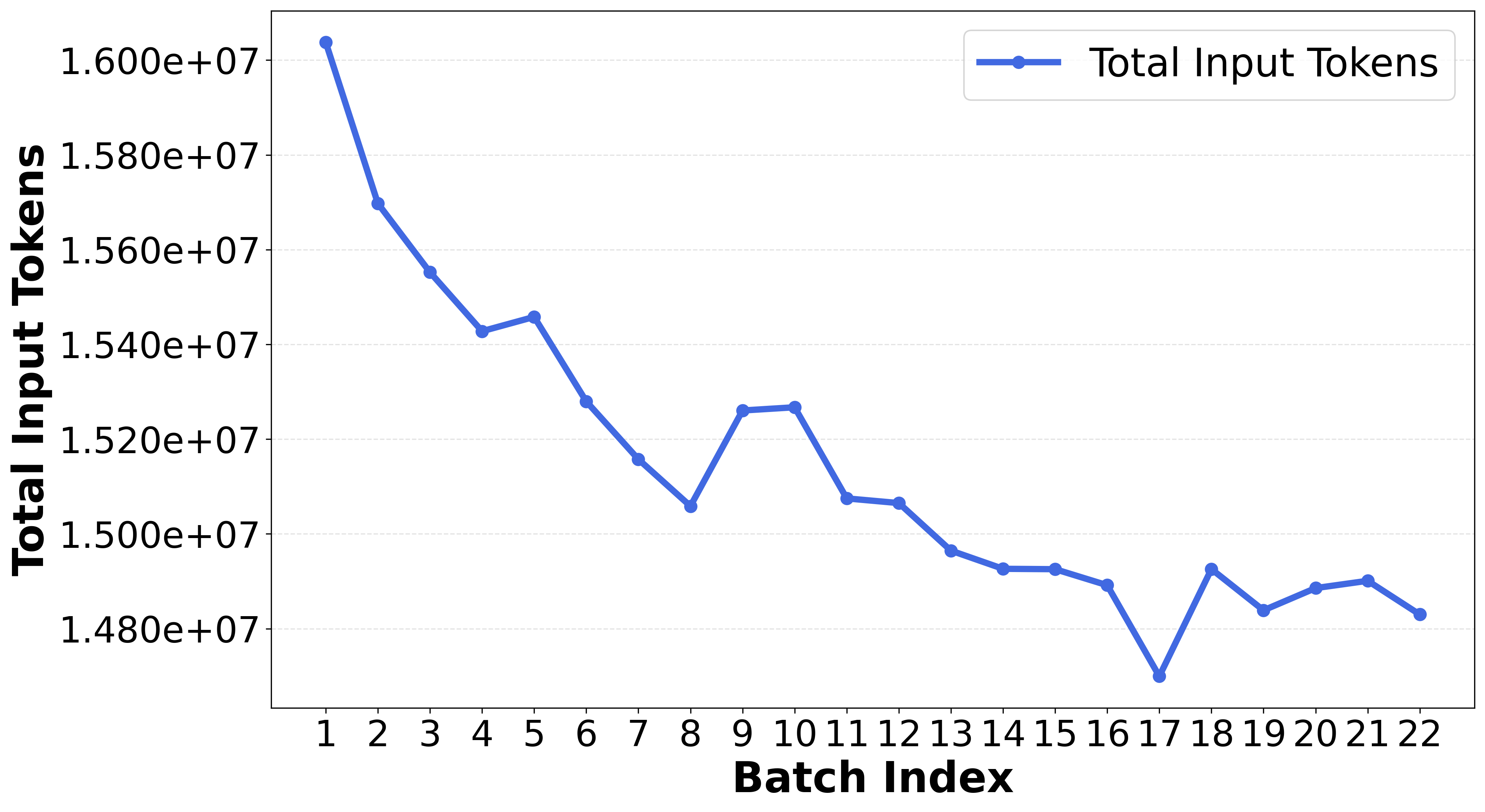}
\caption{Trend of Total Token Consumption across training batches.}
\label{fig:input_tokens_trend}
\end{figure}

\paragraph{System Efficiency} To assess the operational cost of this professional growth, we track the Total Input Tokens processed across all interactions. Analysis reveals a progressive optimization in computational efficiency that directly parallels the clinical improvements. As illustrated in Figure \ref{fig:input_tokens_trend}, the Total Input Tokens consumption per batch declines, dropping from an initial \textbf{16.04} million tokens to \textbf{14.83} million tokens in the final batches. This downward trend is highly significant when paired with the rising clinical metrics: it indicates that through the built-in reflection mechanism, physician agents learn to streamline their diagnostic workflows. By eliminating redundant inquiries and focusing on high-yield clinical data, the agents achieve superior medical outcomes with a reduced computational footprint.

\subsection{Case Studies}

To provide a concrete example of this professional growth and the emergence of collective intelligence, we present two distinct case studies.

Figure \ref{fig:evolution_case} illustrates a specific patient case across the training timeline to demonstrate individual agent evolution. In Batch 1, the physician agent exhibits a scattered diagnostic approach and orders irrelevant tests such as \texttt{NAAT}, \texttt{AST}, and \texttt{\textit{Neisseria gonorrhoeae} culture}, which results in low Examination Precision. This unfocused data gathering ultimately leads to a misdiagnosis of \texttt{gonococcal pharyngitis} and a subsequently ineffective treatment plan with a low alignment score. Conversely, by Batch 22, the same agent demonstrates highly refined clinical judgment. It bypasses redundant tests and directly orders the targeted \texttt{syphilis serology}, \texttt{throat culture}, and \texttt{EBV antibody panel} to achieve high Examination Precision and accurately diagnose the condition as a \texttt{viral infection}. Accompanying this precise diagnosis, the agent formulates a treatment plan that shows a significant improvement in clinical alignment. This qualitative shift perfectly mirrors the quantitative improvements, demonstrating how the high-fidelity patient environment effectively catalyzes the professional evolution of the physician agents.

\begin{figure}[ht]
\includegraphics[width=\columnwidth]{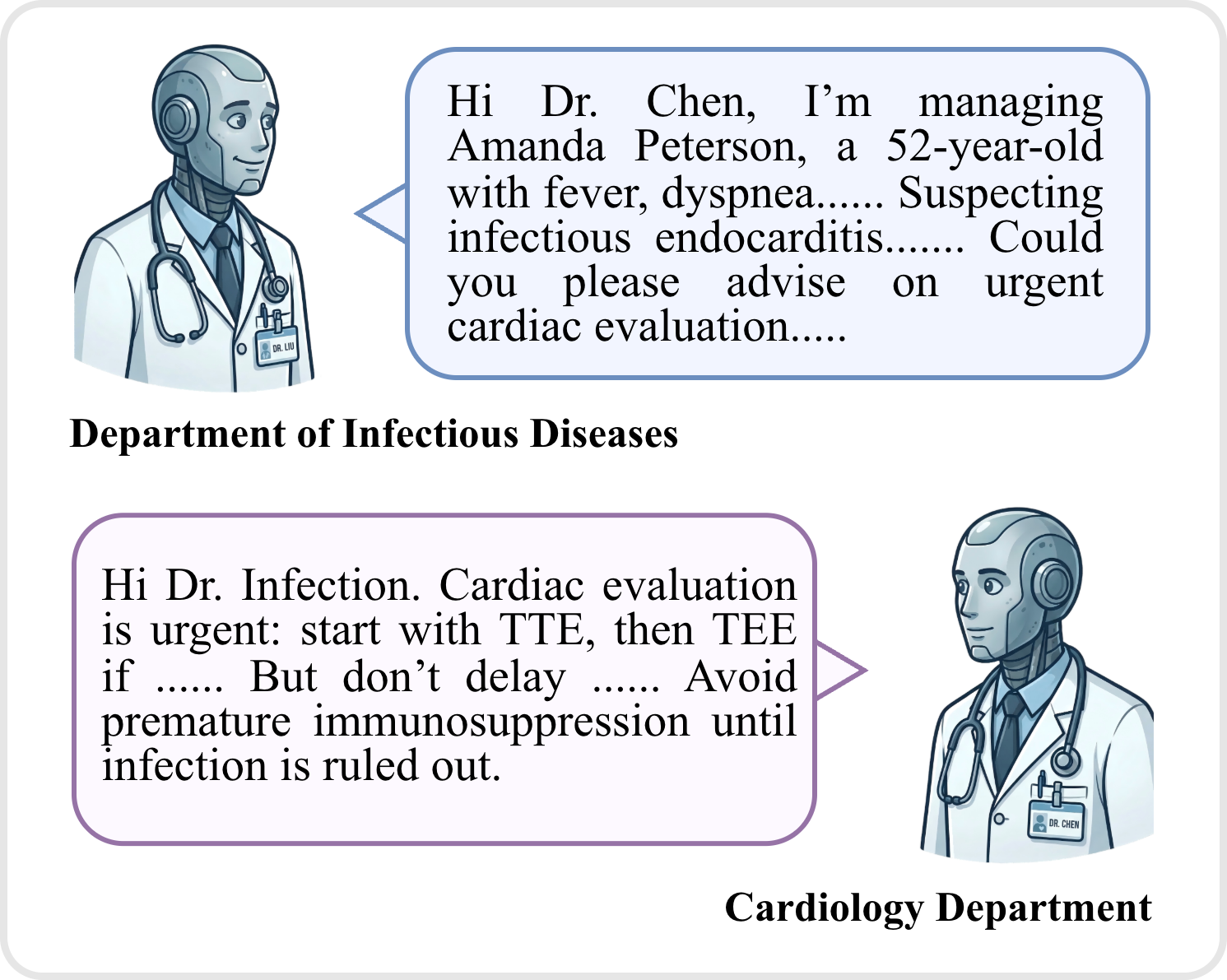}
\caption{An example of cross-departmental physician consultation.}
\label{fig:doctor_consultation}
\end{figure}

Furthermore, analysis of the diagnostic process reveals the spontaneous emergence of sophisticated cooperative behaviors, highlighting the system's collective intelligence. Figure \ref{fig:doctor_consultation} presents a complex scenario involving a patient, Amanda Peterson, presenting with comorbidities that necessitate cross-departmental collaboration. Initially, the physician agent from the Department of Infectious Diseases identifies fever and dyspnea, correctly suspecting infectious endocarditis. Recognizing the limits of its single-domain knowledge, the agent proactively initiates a consultation with the Cardiology Department for urgent cardiac evaluation. The cardiology agent responds with targeted examination recommendations—prioritizing a Transthoracic Echocardiogram (TTE) followed by a Transesophageal Echocardiogram (TEE)—and provides crucial clinical advice regarding subsequent treatment precautions. This interaction highlights the collaborative necessity intrinsic to OpenHospital, demonstrating how agents successfully integrate diverse medical knowledge and debate treatment options to manage complex cases. Instead of relying on isolated reasoning, the agents iteratively synthesize a safe, comprehensive clinical pathway, validating the environment as a training ground for genuine collective intelligence.

\section{Conclusion}

OpenHospital serves as a dedicated arena for the emergence and development of LLM-based Collective Intelligence (CI). By enabling autonomous physician-patient interactions, the platform facilitates the continuous professional evolution of physician agents through a data-in-agent-self paradigm. Furthermore, it establishes a dual-dimensional assessment framework that evaluates both medical proficiency and system efficiency, providing a comprehensive quantitative measure of CI in complex tasks. Our findings validate OpenHospital’s dual role as both a potent evolutionary training ground for agent growth and a rigorous benchmark for quantifying CI in multi-agent systems. Future work will focus on integrating multimodal capabilities and modeling fine-grained temporal disease progression.

\section*{Limitations}
\paragraph{Unimodal Constraints and Information Loss} A primary limitation of the current OpenHospital is its reliance on a text-only modality. While textual records (e.g., chief complaints, medical history, and lab reports) constitute a significant portion of Electronic Health Records (EHRs), real-world clinical diagnosis is inherently multimodal. Our work currently lacks the capability to process medical imaging (such as CT scans, MRIs, or X-rays) or continuous biosignals (e.g., ECG waveforms). Consequently, this restricts the simulation's fidelity in specialties heavily dependent on visual data, such as radiology and dermatology. Future iterations will aim to integrate Vision-Language Models (VLMs) to bridge this gap.

\paragraph{Simplified clinical abstraction} Although OpenHospital is designed to approximate realistic clinical workflows—including multi-stage consultation, testing, and collaborative decision-making—it remains an abstraction of real-world practice. In particular, the benchmark does not model fine-grained temporal progression of disease, evolving symptoms over time, or dynamically changing clinical conditions. As a result, while OpenHospital reflects key elements of diagnostic reasoning and team coordination, it does not fully reproduce the temporal and operational richness of real clinical care.

\section*{Ethical Considerations}

\paragraph{Privacy and Use of Synthetic Data} To strictly uphold data privacy standards, OpenHospital is constructed entirely using synthetic data. We ensure that all patient profiles, medical histories, and dialogue interactions are generatively synthesized, devoid of any real-world patient records or Protected Health Information (PHI). This design choice eliminates the risk of data leakage, enabling the OpenHospital to be openly distributed in full compliance with major privacy regulations, including the Health Insurance Portability and Accountability Act (HIPAA) and the General Data Protection Regulation (GDPR).

\paragraph{Clinical Safety and Research Scope} It is imperative to clarify that OpenHospital is a research environment designed for evolving and benchmarking multi-agent systems collective intelligence, not for clinical decision support. We urge future users of this arena to maintain a clear distinction between simulation results and actionable clinical guidelines, and results obtained within the arena should not be interpreted as validated clinical recommendations.



\bibliography{custom}




\end{document}